\documentclass[runningheads]{llncs}
\usepackage{graphicx}
\usepackage{caption}
\usepackage{subcaption}
\usepackage{multirow}
\usepackage{url}
\usepackage{xcolor}
\usepackage{adjustbox}
\usepackage{float}
\usepackage{xurl}

\begin{document}
\title{Facilitating the Production of Well-tailored Video Summaries for Sharing on Social Media\thanks{This work was supported by the EU Horizon 2020 programme under grant agreement H2020-951911 AI4Media.}}
\iftrue
    \author{Evlampios Apostolidis\inst{}\orcidID{0000-0001-5376-7158} \and
    Konstantinos Apostolidis\inst{}\orcidID{0000-0002-9470-6332} \and
    Vasileios Mezaris\inst{}\orcidID{0000-0002-0121-4364}}
    \authorrunning{E. Apostolidis et al.}
    \institute{Information Technologies Institute - Centre for Research and Technologies, Hellas, 6th km Charilaou - Thermi Road, 57001, Thessaloniki, Greece\\
    \email{\{apostolid,kapost,bmezaris\}@iti.gr}}
\else
    \author{\textit{Paper anonymized for double-blind review}}
    \authorrunning{}
    \institute{\email{}}
\fi

\maketitle 
\begin{abstract}
This paper presents a web-based tool that facilitates the production of tailored summaries for online sharing on social media. Through an interactive user interface, it supports a ``one-click'' video summarization process. Based on the integrated AI models for video summarization and aspect ratio transformation, it facilitates the generation of multiple summaries of a full-length video according to the needs of target platforms with regard to the video's length and aspect ratio.

\end{abstract}
\section{Introduction}

Social media users crave short videos that attract the viewers' attention and can be ingested quickly. Therefore, for sharing on social media platforms, video creators often need a trimmed-down version of their original full-length video. However, different platforms impose different restrictions on the duration and aspect ratio of the video that they accept, e.g., on Facebook's feed videos up to 2 min. appear in a 16:9 ratio, whereas Instagram and Facebook stories usually allow for 20 sec. and are shown in a 9:16 ratio. This makes the generation of tailored versions of video content for sharing on multiple platforms a tedious task. In this paper, we introduce a web-based tool that harnesses the power of AI (Artificial Intelligence) to automatically generate video summaries that encapsulate the flow of the story and the essential parts of the full-length video and are already adapted to the needs of different social media platforms in terms of video length and aspect ratio.

\section{Related Work}

Several video summarization tools can be found online, that are based on AI models. However, most of them produce a textual summary of the video, by analyzing the available \cite{summarize_tech,videohighlight} or automatically-extracted transcripts \cite{mindgrasp,brevify} using NLP (Natural Language Processing) models. Similar solutions are currently provided by various browser plugins \cite{eightify,vidsummize}. Going one step further, another tool creates a video summary by stitching in chronological order the parts of the video that correspond to the selected transcripts for inclusion in the textual summary \cite{pictory}. Focusing on the visual content, Cloudinary released a tool that allows users to upload a video and receive a summarized version of it based on a user-defined summary length (max 30 sec.) \cite{cloudinary}. In addition, Cognitive Mill released a paid AI-based platform which, among other media content management tasks, supports the semi-automatic production of movie trailers and video summaries \cite{cognitivemill}. The proposed solution in this paper is most closely related to the tool of Cloudinary. However, contrary to this tool, our solution offers various options for the video summary duration and applies video aspect ratio transformation techniques to fully meet the specifications of the target video-sharing platform.

\section{Proposed Solution}

The proposed solution (available at \textcolor{blue}{\url{https://idt.iti.gr/summarizer}}) is an extension of the web-based service for video summarization, presented in \cite{collyda2020web}. It is composed of a front-end user interface (UI) that allows interaction with the user (presented in Sec. \ref{subsec:frontend}), and a back-end component that analyses the video and produces the video summary (discussed in Sec. \ref{subsec:backend}). The front-end and back-end communication is carried out via REST calls that initiate the analysis, periodically request its status, and, after completion, retrieve the analysis results (i.e., the video summary) for presentation to the user. Our solution extends \cite{collyda2020web} by: i) using an advanced AI-based method for video summarization, ii) integrating an AI-based approach for spatially cropping the video given a target aspect ratio, and iii) supporting customized values for the target duration and aspect ratio of the generated video summary.

\subsection{Front-end UI}
\label{subsec:frontend}

\begin{figure}[t!]
    \centering
    \begin{subfigure}{\linewidth}
         \centering
        \frame{\includegraphics[width=\linewidth]{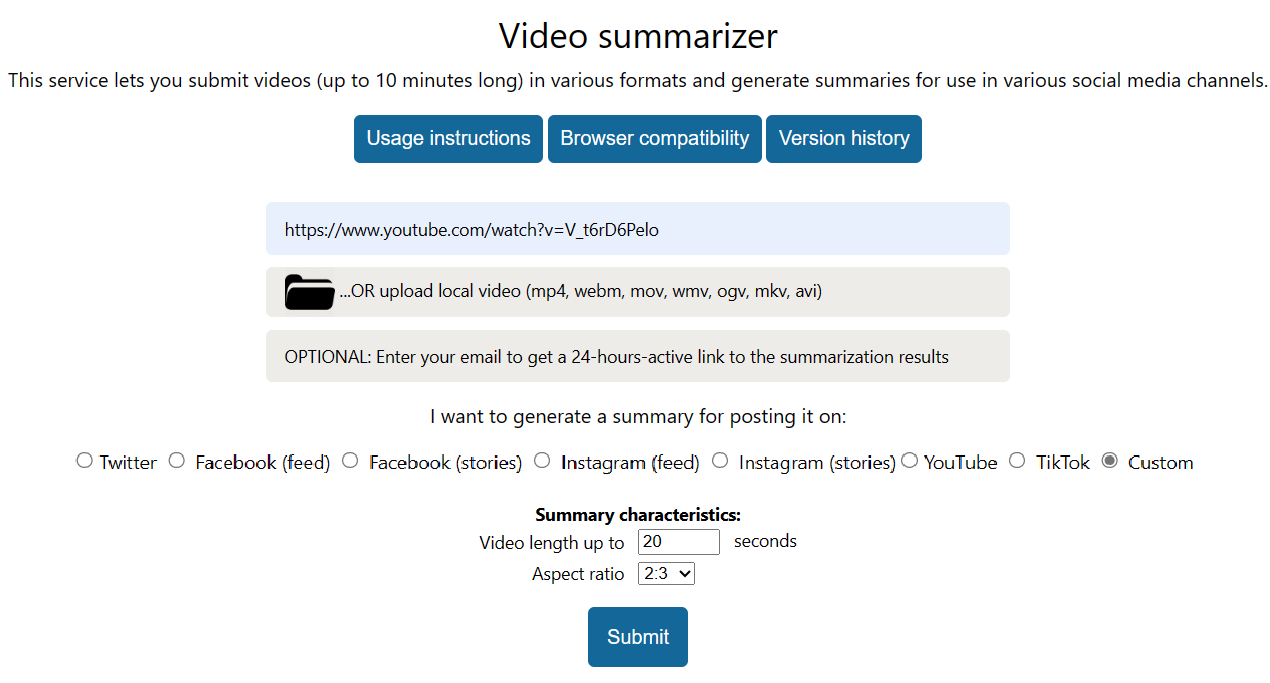}}
        \caption{The landing page of the UI.}
    \end{subfigure}
    
    \begin{subfigure}{\linewidth}
        \centering
        \frame{\includegraphics[width=\linewidth]{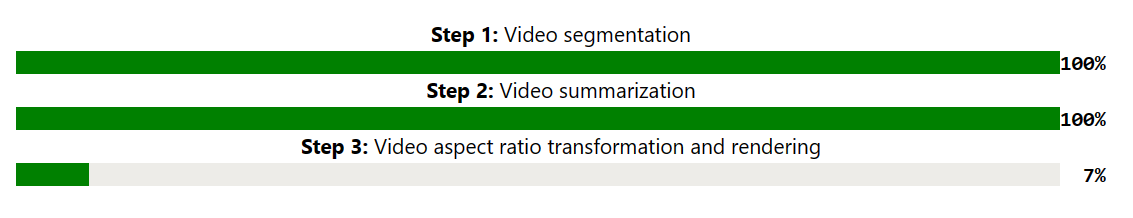}}
        \caption{The progress-reporting bars.}
    \end{subfigure}
    
    \begin{subfigure}{\linewidth}
        \centering
        \frame{\includegraphics[width=\linewidth]{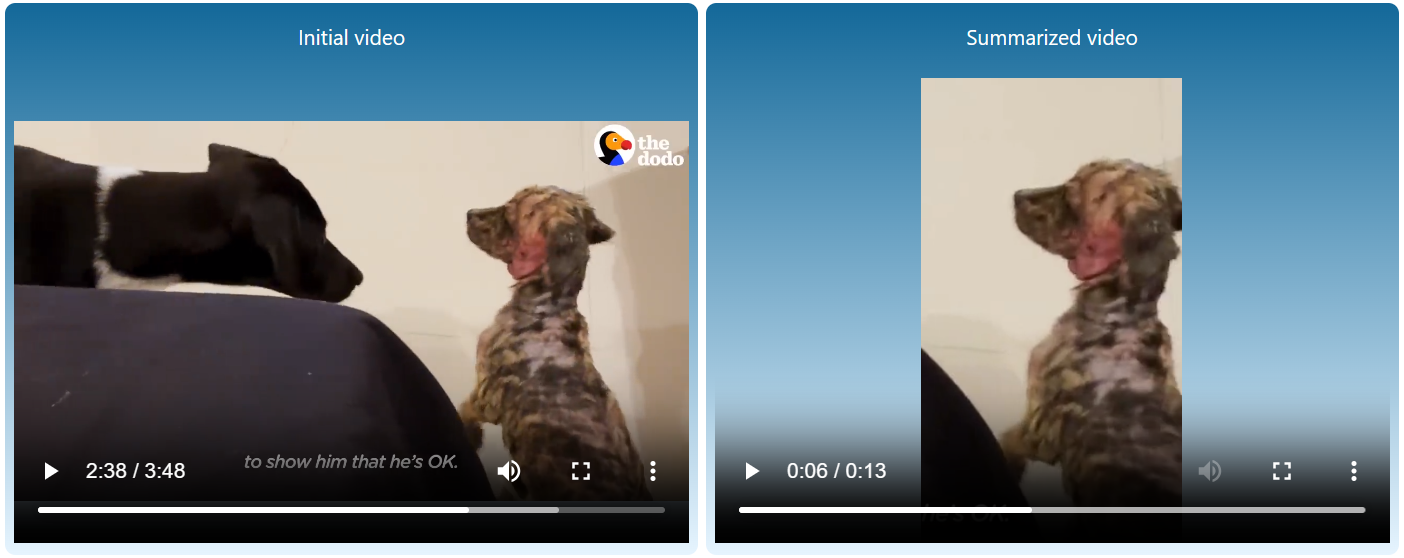}}
        \caption{The video players of the page showing the analysis results.}
    \end{subfigure}
    \caption{Instances of the updated and extended UI.}
    \label{fig:screens}
\end{figure}

The UI of the proposed solution (see the top part of Fig.~\ref{fig:screens}) allows the user to submit a video (that is either online-available or locally-stored in the user's device) for summarization, and choose the duration and aspect ratio of the produced summary. This choice can be made either by selecting among presets for various social media channels, or in a fully-custom manner. After initiating the analysis, the user can monitor its progress (see the middle part of Fig.~\ref{fig:screens}) and submit additional requests while the previous ones are being analyzed. When the analysis is completed, the original video and the produced summary are shown to the user through an interactive page containing two video players that support all standard functionalities (see the bottom part of Fig.~\ref{fig:screens}); through the same page, the user is able to download the produced video summary. Further details about the supported online sources, the permitted file types, and the management of the submitted and produced data, can be found in \cite{collyda2020web}.

\begin{table}[t]
\centering
\caption{Performance (F-Score (\%)) of SUM-GAN-AEE and AC-SUM-GAN on SumMe and TVSum; the last row reports AC-SUM-GAN's performance for augmented training data.}
\label{tab:sum_results_1}
\begin{tabular}{|l|c|c|} 
        \hline
        Method & SumMe & TVSum \\ \hline \hline
        SUM-GAN-AAE~\cite{apostolidis2020unsupervised} (used in \cite{collyda2020web})   & 48.9  & 58.3  \\ \hline
        AC-SUM-GAN~\cite{9259058} & 50.8 & 60.6 \\ \hline \hline
        AC-SUM-GAN$_{aug}$ (used now) & \textbf{52.0}  & \textbf{61.0}  \\ \hline
\end{tabular}
\end{table}

\begin{table}[t]
    \caption{Used datasets for training and evaluating the AC-SUM-GAN model.}
    \label{tab:datasets}
    \centering
    \begin{tabular}{|l|c|c|c|} 
        \hline
        Dataset & Videos & Duration (min) & Content \\ \hline \hline
        SumMe~\cite{gygli2014creating} & 25 & 1-6 & \begin{tabular}[c]{@{}c@{}}holidays, events, sports, \\ travelling, cooking\end{tabular} \\ \hline
        TVSum~\cite{7299154} & 50 & 1 - 11 & \begin{tabular}[c]{@{}c@{}}news, how-to's, user-generated, \\ documentaries\end{tabular} \\ \hline    
        OVP~\cite{de2011vsumm}     & 50       & 1 - 4          & \begin{tabular}[c]{@{}c@{}}documentary, educational,\\  ephemeral, historical, lecture\end{tabular} \\ \hline
        YouTube~\cite{de2011vsumm} & 50       & 1 - 10         & \begin{tabular}[c]{@{}c@{}}cartoons, sports, tv-shows, \\ commercial, home videos\end{tabular} \\ \hline
    \end{tabular}
\end{table}

\subsection{Back-end component}
\label{subsec:backend}

The submitted video is initially fragmented to shots using a pre-trained model of the method from \cite{souvcek2020transnet}, which exhibits $11\%$ improved performance on the RAI dataset \cite{10.1007/978-3-319-23192-1_67} compared to the previous (used in \cite{collyda2020web}) approach. Following, \textbf{video summarization} is performed using a pre-trained model of AC-SUM-GAN \cite{9259058}, a top-performing unsupervised video summarization method \cite{vs_survey}. This method embeds an Actor-Critic model into a Generative Adversarial Network and formulates the selection of important video fragments as a sequence generation task. At training time, the Actor-Critic model utilizes the Discriminator's feedback as a reward, to progressively explore a space of actions and learn a value function (Critic) and a policy (Actor) for key-fragment selection. As shown in Table \ref{tab:sum_results_1}, AC-SUM-GAN performs much better than SUM-GAN-AAE \cite{apostolidis2020unsupervised} (used in \cite{collyda2020web}), on the SumMe \cite{gygli2014creating} and TVSum \cite{7299154} benchmark datasets for video summarization. Both methods learn the task with the help of a summary-to-video reconstruction mechanism and using the received feedback from an adversarially-trained Discriminator. We argue that the advanced performance of AC-SUM-GAN relates to the use of this feedback as a reward for training an Actor-Critic model and learning a good policy for key-fragment selection, rather than using it as part of a loss function to train a bi-directional LSTM for frame importance estimation. 

The proposed solution uses a model of AC-SUM-GAN that has been trained using augmented data. Following the typical approach in the literature \cite{vs_survey}, we extended the pool of training samples of the SumMe and TVSum datasets, by including videos of the OVP and YouTube \cite{de2011vsumm} datasets. As presented in Table \ref{tab:datasets}, the utilized data include videos from various categories and thus facilitate the training of a general-purpose video summarization model. Nevertheless, we anticipate a better summarization performance on videos from the different categories found in the used datasets, such as tutorials, ``how-to'', product demos, and event videos (e.g., birthday parties) that are commonly shared on social media, compared to the expected performance on videos from completely unseen categories, such as movies, football games and music shows. This data augmentation process resulted in improvements on both benchmarking datasets (see the last row of Table \ref{tab:sum_results_1}) and to a very competitive performance against several state-of-the-art (SotA) unsupervised methods from the literature that have been assessed under the same evaluation settings (see Table \ref{tab:sum_results_2}).

\begin{table}[t]
\centering
\caption{Performance comparison (F-Score (\%)) with SotA unsupervised approaches after using augmented training data. The reported scores for the listed methods are from the corresponding papers.}
\label{tab:sum_results_2}
\begin{tabular}{|l|c|c|} 
        \hline
        Method & SumMe & TVSum \\ \hline \hline
        ACGAN~\cite{He:2019:UVS:3343031.3351056} & 47.0 & 58.9 \\ \hline
        RSGN$_{unsup}$~\cite{9399800} & 43.6 & 59.1 \\ \hline
        3DST-UNet~\cite{10.1109/TIP.2022.3143699} & 49.5 & 58.4 \\ \hline
        DSR-RL-GRU~\cite{9428142} & 48.5 & 59.2 \\ \hline
        ST-LSTM~\cite{10.1007/s11042-022-12901-4} & \textbf{52.0} & 58.1 \\ \hline
        CAAN~\cite{LIANG2022108840} & 50.9 & 59.8 \\ \hline
        SUM-GDA$_{unsup}$~\cite{LI2021107677} & 50.2 & 60.5 \\ \hline
        SUM-FCN$_{unsup}$~\cite{10.1007/978-3-030-01258-8_22} & 51.1 & 59.2 \\ \hline \hline
        AC-SUM-GAN$_{aug}$ & \textbf{52.0} & \textbf{61.0} \\ \hline
\end{tabular}
\end{table}

\begin{table}[t]
    \caption{Video aspect ratio transformation performance (IoU (\%)) on the RetargetVid dataset.}
    \label{tab:sc_results}
    \centering
    \begin{tabular}{|l|l|c|c|c|} 
        \hline
        & Method & Worst & Best & Mean  \\ \hline \hline
        \multirow{2}{*}{\begin{tabular}[c]{@{}l@{}} 1:3 target\\ aspect ratio\end{tabular}} & SVC (used in \cite{apostolidis2021web}) & 51.7 & 53.8 & 52.9 \\ \cline{2-5} 
        & SVC$_{ext}$ (used now) & \textbf{53.8} & \textbf{57.6} & \textbf{55.6} \\ \hline \hline

        \multirow{2}{*}{\begin{tabular}[c]{@{}l@{}} 3:1 target\\ aspect ratio\end{tabular}} & SVC (used in \cite{apostolidis2021web}) & 74.4 & 77.0 & 75.3  \\ \cline{2-5} 
        & SVC$_{ext}$ (used now) & \textbf{76.3} & \textbf{78.0} & \textbf{77.6} \\ \hline
    \end{tabular}
\end{table}

To minimize the possibility of losing semantically-important visual content or resulting in visually-unpleasant results during \textbf{video aspect ratio transformation} (that would be highly possible when using naive approaches, such as fixed cropping of a central area of the video frames, or padding of black borders to reach the target aspect ratio) the proposed solution integrates an extension of the smart video cropping (SVC) method of \cite{apostolidis2021web}. The latter starts by computing the saliency map for each frame that was chosen for inclusion in the video summary. Then, to select the main part of the viewers' focus, the integrated method applies a filtering-through-clustering procedure on the pixel values of each predicted saliency map. Finally, it infers a single point as the center of the viewer's attention and computes a crop window for each frame based on the displacement of this point. The applied extension on \cite{apostolidis2021web}, relates to the use of a SotA method for saliency prediction \cite{hu2023tinyhd}, that resulted in improved performance on the RetargetVid dataset \cite{icip_smartvidcrop}. As shown in Table \ref{tab:sc_results}, the averaged Intersection-over-Union (IoU) scores for all video frames have been increased by over 2 percentage points.

\section{Conclusions}
\label{sec:conclusions}
In this paper, we presented a web-based tool that facilitates the generation of video summaries that are tailored to the specifications of various social media platforms, in terms of video length and aspect ratio. After reporting on the applied extensions to a previous instance of the tool, we provided more details about the front-end user interface and the back-end component of this technology, and we documented the advanced performance of the newly integrated AI models for video summarization and aspect ratio transformation. This tool will be demonstrated in MMM2024, while the participants in the relevant demonstration session will have the opportunity to test our tool in real-time.

%
%
%

\begin{thebibliography}{10}
\providecommand{\url}[1]{\texttt{#1}}
\providecommand{\urlprefix}{URL }
\providecommand{\doi}[1]{https://doi.org/#1}

\bibitem{brevify}
{Brevify: Video Summarizer}. \url{https://devpost.com/software/brevify-video-summarizer}, accessed: 2023-09-29

\bibitem{cloudinary}
{Cloudinary: Easily create engaging video summaries}. \url{https://smart-ai-transformations.cloudinary.com}, accessed: 2023-09-29

\bibitem{cognitivemill}
{Cognitive Mill: Cognitive Computing Cloud Platform For Media And Entertainment}. \url{https://cognitivemill.com}, accessed: 2023-09-29

\bibitem{eightify}
{Eightify: Youtube Summary with ChatGPT}. \url{https://chrome.google.com/webstore/detail/eightify-youtube-summary/cdcpabkolgalpgeingbdcebojebfelgb}, accessed: 2023-09-29

\bibitem{pictory}
Pictory: Automatically summarize long videos. \url{https://pictory.ai/pictory-features/auto-summarize-long-videos}, accessed: 2023-09-29

\bibitem{summarize_tech}
summarize.tech: {AI}-powered video summaries. \url{https://www.summarize.tech}, accessed: 2023-09-29

\bibitem{videohighlight}
{Video Highlight}: the fastest way to summarize and take notes from videos. \url{https://videohighlight.com}, accessed: 2023-09-29

\bibitem{mindgrasp}
{Video Summarizer - Summarize Youtube Videos}. \url{https://mindgrasp.ai/video-summarizer}, accessed: 2023-09-29

\bibitem{vidsummize}
{VidSummize - AI YouTube Summary with Chat GPT}. \url{https://chrome.google.com/webstore/detail/vidsummize-ai-youtube-sum/gidcfccogfdmkfdfmhfdmfnibafoopic}, accessed: 2023-09-29

\bibitem{apostolidis2020unsupervised}
Apostolidis, E., Adamantidou, E., Metsai, A.I., Mezaris, V., Patras, I.: Unsupervised video summarization via attention-driven adversarial learning. In: Proc. 26th Int. Conf. MultiMedia Modeling (MMM), Part I 26. pp. 492--504. Springer (2020)

\bibitem{9259058}
Apostolidis, E., Adamantidou, E., Metsai, A.I., Mezaris, V., Patras, I.: {AC-SUM-GAN}: Connecting actor-critic and generative adversarial networks for unsupervised video summarization. IEEE Trans. on Circuits and Systems for Video Technology  \textbf{31}(8),  3278--3292 (2021). \doi{10.1109/TCSVT.2020.3037883}

\bibitem{vs_survey}
Apostolidis, E., Adamantidou, E., Metsai, A.I., Mezaris, V., Patras, I.: Video summarization using deep neural networks: A survey. Proceedings of the IEEE  \textbf{109}(11),  1838--1863 (2021). \doi{10.1109/JPROC.2021.3117472}

\bibitem{icip_smartvidcrop}
Apostolidis, K., Mezaris, V.: A fast smart-cropping method and dataset for video retargeting. In: Proc. IEEE Int. Conf. on Image Processing (ICIP). pp. 1956--1960 (2021)

\bibitem{apostolidis2021web}
Apostolidis, K., Mezaris, V.: A web service for video smart-cropping. In: 2021 IEEE Int. Symposium on Multimedia (ISM). pp. 25--26. IEEE (2021)

\bibitem{10.1007/978-3-319-23192-1_67}
Baraldi, L., Grana, C., Cucchiara, R.: Shot and scene detection via hierarchical clustering for re-using broadcast video. In: Proc. 16th Int. Conf. Computer Analysis of Images and Patterns (CAIP), Part I 16. pp. 801--811. Springer (2015)

\bibitem{collyda2020web}
Collyda, C., Apostolidis, K., Apostolidis, E., Adamantidou, E., Metsai, A.I., Mezaris, V.: A web service for video summarization. In: ACM Int. Conf. on Interactive Media Experiences (IMX). pp. 148--153 (2020)

\bibitem{de2011vsumm}
De~Avila, S.E.F., Lopes, A.P.B., da~Luz~Jr, A., de~Albuquerque~Ara{\'u}jo, A.: {VSUMM: A} mechanism designed to produce static video summaries and a novel evaluation method. Pattern recognition letters  \textbf{32}(1),  56--68 (2011)

\bibitem{gygli2014creating}
Gygli, M., Grabner, H., Riemenschneider, H., Van~Gool, L.: Creating summaries from user videos. In: 13th European Conf. on Computer Vision (ECCV), Zurich, Switzerland, 2014, Proc., Part VII 13. pp. 505--520. Springer (2014)

\bibitem{He:2019:UVS:3343031.3351056}
He, X., Hua, Y., Song, T., Zhang, Z., Xue, Z., Ma, R., Robertson, N., Guan, H.: Unsupervised video summarization with attentive conditional generative adversarial networks. In: Proc. of the 27th ACM Int. Conf. on Multimedia (MM '19). pp. 2296--2304. ACM, New York, NY, USA (2019)

\bibitem{hu2023tinyhd}
Hu, F., Palazzo, S., Salanitri, F.P., Bellitto, G., Moradi, M., Spampinato, C., McGuinness, K.: Tinyhd: Efficient video saliency prediction with heterogeneous decoders using hierarchical maps distillation. In: Proc. of the IEEE/CVF Winter Conf. on Applications of Computer Vision. pp. 2051--2060 (2023)

\bibitem{LI2021107677}
Li, P., Ye, Q., Zhang, L., Yuan, L., Xu, X., Shao, L.: Exploring global diverse attention via pairwise temporal relation for video summarization. Pattern Recognition  \textbf{111},  107677 (2021)

\bibitem{LIANG2022108840}
Liang, G., Lv, Y., Li, S., Zhang, S., Zhang, Y.: Video summarization with a convolutional attentive adversarial network. Pattern Recognition  \textbf{131},  108840 (2022)

\bibitem{10.1109/TIP.2022.3143699}
Liu, T., Meng, Q., Huang, J.J., Vlontzos, A., Rueckert, D., Kainz, B.: Video summarization through reinforcement learning with a 3d spatio-temporal u-net. Trans. Img. Proc.  \textbf{31},  1573–1586 (jan 2022)

\bibitem{10.1007/s11042-022-12901-4}
Min, H., Ruimin, H., Zhongyuan, W., Zixiang, X., Rui, Z.: Spatiotemporal two-stream lstm network for unsupervised video summarization. Multimedia Tools and Applications  \textbf{81},  40489--40510 (2022)

\bibitem{9428142}
Phaphuangwittayakul, A., Guo, Y., Ying, F., Xu, W., Zheng, Z.: Self-attention recurrent summarization network with reinforcement learning for video summarization task. In: Proc. of the 2021 IEEE Int. Conf. on Multimedia and Expo (ICME). pp.~1--6 (2021). \doi{10.1109/ICME51207.2021.9428142}

\bibitem{10.1007/978-3-030-01258-8_22}
Rochan, M., Ye, L., Wang, Y.: {Video summarization using fully convolutional sequence networks}. In: Ferrari, V., Hebert, M., Sminchisescu, C., Weiss, Y. (eds.) Europ. Conf. on Computer Vision (ECCV) 2018. pp. 358--374. Springer International Publishing, Cham (2018)

\bibitem{7299154}
Song, Y., Vallmitjana, J., Stent, A., Jaimes, A.: {TVSum: Summarizing web videos using titles}. In: IEEE Conf. on Computer Vision and Pattern Recognition (CVPR). pp. 5179--5187 (2015). \doi{10.1109/CVPR.2015.7299154}

\bibitem{souvcek2020transnet}
Sou{\v{c}}ek, T., Loko{\v{c}}, J.: {Transnet V2}: an effective deep network architecture for fast shot transition detection. arXiv preprint arXiv:2008.04838  (2020)

\bibitem{9399800}
Zhao, B., Li, H., Lu, X., Li, X.: Reconstructive sequence-graph network for video summarization. IEEE Trans. on Pattern Analysis and Machine Intelligence pp.~1--1 (2021). \doi{10.1109/TPAMI.2021.3072117}

\end{thebibliography}
%

\end{document}